\documentclass[letterpaper, 10 pt, conference]{ieeeconf}  % Comment % \documentclass[lettersize,journal]{IEEEtran}
\usepackage{booktabs}
\usepackage{amsmath,amsfonts}
\usepackage{algorithmic}
\usepackage{algorithm}
\usepackage{array}
\usepackage{textcomp}
\usepackage{xspace}
\usepackage{stfloats}
\usepackage{pifont}    % for \ding{55}
\usepackage{url}
\usepackage{multirow}
\usepackage{lipsum}
\usepackage{graphicx}
\usepackage{makecell} 
\usepackage{verbatim}
\usepackage{graphicx}

\usepackage[sort,compress]{cite}
\makeatletter
\let\NAT@parse\undefined
\makeatother
\usepackage[
   colorlinks=true,
   linkcolor=black,
   citecolor=black,
   urlcolor=black,
   filecolor=black,
   pagebackref=false,
   hypertexnames=false,
   plainpages=false
]{hyperref}

\usepackage{graphicx}
\usepackage{amsmath}
\usepackage{array}
\usepackage{colortbl}
\usepackage[table,xcdraw]{xcolor}
\usepackage{arydshln}
\usepackage[normalem]{ulem}  % for strikethrough text (\sout); normalem keeps the behavior of emph as expected

\newcommand{\redx}{{\color{red}\ding{55}}}
\definecolor{darkgreen}{RGB}{0,128,0}
\newcommand{\greencheck}{{\color{darkgreen}\ding{51}}}
\usepackage{tikz}
\usetikzlibrary{arrows.meta,positioning,backgrounds,fit,calc}
\definecolor{agentA}{HTML}{E76E51} % orange  (Okabe-Ito, colorblind-safe)
\definecolor{agentB}{HTML}{2A9D8E} % blue
\definecolor{bg}{HTML}{9DB0B5}
\definecolor{ink}{HTML}{284651}
\usepackage{amssymb}

\newcommand{\blue}[1]{\textcolor{black}{#1}}
\definecolor{myteal}{HTML}{95CEC7}
\definecolor{myyellow}{HTML}{FAF3B3}
\definecolor{mycoral}{HTML}{F8CA8C}

\NewDocumentCommand{\placeholderfig}{O{4cm} m}{%
  \fbox{%
    \begin{minipage}[c][#1][c]{\dimexpr\linewidth-2\fboxsep-2\fboxrule\relax}
      \centering\small\itshape #2
    \end{minipage}%
  }%
}

\newcommand{\ours}{GaussLite\xspace}
\newcommand{\secref}[1]{Sec.~\ref{#1}}
\newcommand{\figref}[1]{Fig.~\ref{#1}}
\newcommand{\tabref}[1]{Tab.~\ref{#1}}
\newcommand{\eg}{e.g.\@\xspace}

\title{GaussLite: Online Task-Conditioned 3D Gaussian Splatting for \\Real-Time Robotic Mapping}

\author{Annika Thomas, Mason Peterson and Jonathan P. How%
\thanks{The authors are with the Aerospace Controls Laboratory, MIT. Email: \{annikat, jhow\}@mit.edu}}
\begin{document}

\maketitle

\begin{abstract}
Existing 3D Gaussian Splatting (3DGS) systems distribute representation capacity uniformly across a scene, ignoring the fact that many downstream robotic tasks engage only a fraction of the reconstructed geometry. This causes valuable onboard compute to be allocated towards optimizing irrelevant parts of the scene, either limiting online capacity or under-optimizing the most relevant parts of the scene. We introduce \ours, a task-driven 3DGS mapping system that conditions its representation density on a natural-language task specification. Given a posed RGB-D stream and a task such as ``\textit{prepare to pick up the object on the desk},'' \ours uses a one-shot LLM parser to extract target and anchor objects, which are grounded per-frame by an open-vocabulary detector and segmented to produce per-pixel relevance masks in real time. The mapper allocates seeding density, gradient flow and scaling by task relevance. At matched Gaussian budget and real-time mapping at 4 Hz on resource-constrained hardware, \ours outperforms baselines on ROI PSNR on the Replica Dataset by an average $+2.72$ dB and on a real-hardware demonstration in indoor and outdoor settings by $+2.23$ dB.  We further show that two task-specialized agents' maps can be fused into a single shared map via per-voxel voting on active-optimization counts in real time, outperforming concatenation by $+3.42$ dB while only sharing an average $7.08\%$ of the map.
\vspace{0.3cm}
\end{abstract}

\section{Introduction}

% Field anchor: online 3DGS on robot hardware
3D Gaussian splatting (3DGS) has emerged as an expressive scene representation for photorealistic rendering~\cite{kerbl2023gaussiansplatting}. A growing line of online Gaussian mapping systems~\cite{yugay2024gaussianslam,hu2025splatmap,li2025densesplat} integrate 3DGS into robotic SLAM due to the highly accurate maps and trajectory estimates this technology produces.  On robot hardware, the per-frame mapping loop must run in real time, memory is bounded, and the compute budget is typically too tight to refine the full scene at uniform fidelity, yet current online GS-SLAM systems still allocate the representation uniformly over the visible scene, spending unnecessary onboard compute on regions that could be irrelevant to the task at hand. These constraints compound across robot teams, where the inter-agent link is itself a bounded resource that cannot support an entire reconstruction.

\begin{figure}[t]
\centering
\includegraphics[width=1.0\linewidth, trim = {3cm 0 3cm 0}, clip]{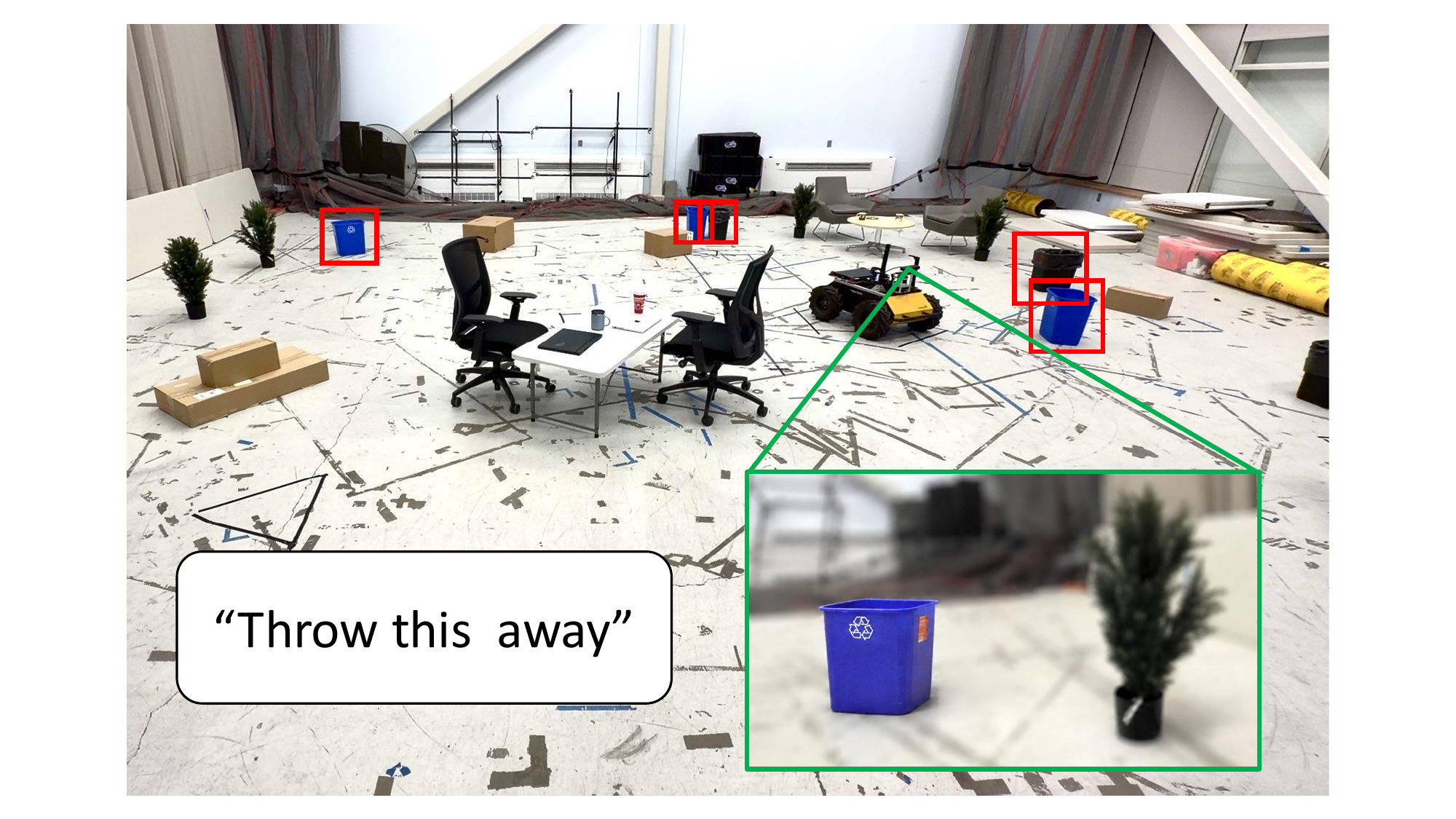}
\caption{Given open-set natural language inputs, GaussLite concentrates Gaussian density and optimization budget on task-relevant regions to achieve higher rendering quality on relevant objects while mapping scenes in real time (e.g. on trash cans given a task ``throw this away'').}
\label{fig:teaser}
\vspace*{-12pt}
\end{figure}

% The gap: everyone reallocates by geometry, no one by task
Prior work has addressed the cost of 3DGS from three directions. Post-hoc compression and pruning methods include global significance pruning~\cite{fan2024lightgaussian}, principled sensitivity pruning~\cite{hanson2024pup3dgs}, minimum-count representations with local distinctiveness~\cite{lee2025omg3dgs}, and deployment-aware compression control~\cite{zhang2025controlgs}. Level-of-detail methods, including Hierarchical-3DGS~\cite{kerbl2024hierarchical}, LODGE~\cite{kulhanek2025lodge}, CLoD-GS~\cite{cheng2026clodgs}, and Octree-GS~\cite{ren2024octreegs}, select Gaussian subsets by camera distance to support efficient rendering of large scenes, but the underlying maps remain large and memory-intensive. In the SLAM setting, online mapping
systems~\cite{hu2025splatmap,li2025densesplat,deng2024compactgsslam} adapt densification to evolving geometry and tracking uncertainty. All of these methods reduce model size, per-view rendering cost, or per-frame mapping cost, but the allocation criteria used are geometric, whether global significance, camera distance, local redundancy, or photometric error, rather than task relevance. As a result, a robot instructed to water plants near a wall may allocate substantial representational capacity to irrelevant regions, such as the ceiling, instead of concentrating fidelity where the task is actually performed.

% Insight: biological vision is non-uniform, online 3DGS should be too
Biological vision is non-uniform by design. The fovea encodes a narrow high-acuity region while the periphery provides a coarse scaffold, and the observer's task determines where the eye fixates~\cite{hayhoe2005eye,yarbus1967eye,land2001roles}. We propose that this principle should transfer to online 3D Gaussian splatting, with representation capacity following the task rather than the geometry. Recent works have established the importance of task-relevant rendering quality: active view selection~\cite{jiang2024fisherrf,pan2022activenerf} chooses where to look, VR foveated rendering~\cite{franke2025vrsplatting,kaplanyan2019deepfovea,lin2025rtgs} adapts rendering to gaze, VISTA~\cite{nagami2026vista} steers trajectories toward task-relevant objects, and GaussianLens~\cite{weng2025gaussianlens} refines user-selected regions of a pre-built 3DGS map. \ours{} is the first to accept natural-language task inputs and spatially allocate the scene representation online (\figref{fig:teaser}). We are also the first to extend this principle to robot teams, where agents pursuing different tasks specialize on different task-relevant regions and exchange only the small, high-confidence subset of each map that the task invested in.

% Mechanics
\ours{} converts a posed RGB-D stream and a natural-language task description into a task-conditioned Gaussian map through three new components. A task-to-attention front-end grounds the natural-language task in each keyframe by combining structured LLM parsing, open-vocabulary detection, and promptable segmentation with a 3D spatial filter that prunes detections inconsistent with the robot's own evolving map, producing a per-pixel relevance mask. A relevance-driven mapping loop propagates this mask into the representation, where each Gaussian carries a relevance score and an active-optimization counter, and seeding density, gradient flow, and densification are allocated by relevance. A multi-agent fusion stage merges task-specialized maps from multiple robots without global re-optimization: a per-voxel vote on each Gaussian's active-optimization counter selects the better-trained source per region, so agents only exchange the small task-specialized subset of their maps that carries meaningful information.
% --- Contributions ---
The resulting contributions of this work are:
\begin{enumerate}
\item \textbf{An open-set task-to-attention pipeline for 3DGS mapping.} Structured LLM parsing, open-vocabulary detection, promptable segmentation, and 3D spatial-relation filtering jointly convert a natural-language task into per-frame pixel-level relevance masks within the per-frame budget of an online mapper (\secref{sec:attention}).

\item \textbf{Relevance-driven Gaussian mapping.} The mapper allocates seeding density, gradient flow, and initial scale by relevance, with an average $+$2.72 dB ROI PSNR improvement on Replica and $+$2.23 dB ROI PSNR improvement on real-world Campus scene/task pairs against three baselines at matched Gaussian budget and real-time mapping (\secref{sec:results}). 
% \todo{emphasize why these numbers are a good contirbution - in context of other developments; maybe a range of improvements}.

\item \textbf{Task-specialized multi-agent fusion.} A per-voxel vote on each Gaussian's active-optimization count fuses task-specialized agents into a single shared map while exchanging only the task-specialized subset of each map, outperforming naive concatenation of baselines by an average $+$3.42 dB ROI PSNR while only sharing on average 7.08$\%$ of the map (\secref{sec:fusion}).
\end{enumerate}

\section{Related Work}
\label{sec:related}

\subsection{Gaussian-splatting SLAM}
3D Gaussian splatting has been integrated into visual SLAM using both monocular~\cite{matsuki2024monogs} and RGB-D sensing
modalities~\cite{yugay2024gaussianslam,keetha2024splatam}, providing real-time rendering and high-fidelity mapping that implicit representations such as NICE-SLAM~\cite{zhu2022niceslam} cannot match at comparable frame rates.  Scaling these systems to large or multi-agent settings introduces additional challenges:
GRAND-SLAM~\cite{thomas2025grandslam} addresses multi-agent consistency through submap-based local optimization and inter-robot loop closure. SplatMAP~\cite{hu2025splatmap} and DenseSplat~\cite{li2025densesplat} adapt densification to SLAM's evolving dense geometry and to sparse-view gap filling, respectively, and Compact 3D Gaussian SLAM~\cite{deng2024compactgsslam} reduces the redundancy inherent in SLAM-generated Gaussians via sliding-window masking and codebook quantization. These systems have accelerated mapping computation, but the underlying Gaussian representation remains allocated uniformly across the visible scene, with no mechanism to prioritize regions that are relevant to the robot's task.

\subsection{Compression, pruning, and level-of-detail for 3DGS}
A substantial body of work reduces the cost of a trained 3DGS model through post-hoc compression and pruning. LightGaussian~\cite{fan2024lightgaussian} identifies Gaussians with minimal global significance and applies pruning, knowledge distillation, and vector quantization to achieve 15$\times$ compression.  PUP 3D-GS~\cite{hanson2024pup3dgs} computes a principled second-order sensitivity score and prunes 90\% of Gaussians while preserving
foreground detail.  Optimized Minimal
3DGS~\cite{lee2025omg3dgs} combines a local distinctiveness metric with compact attribute representations to produce sub-5\,MB scene models, and ControlGS~\cite{zhang2025controlgs} maps a single global hyperparameter to the Gaussian-count-vs-quality curve for deployment-aware control.  A parallel line of work constructs multi-scale or level-of-detail
representations.  
% In the NeRF regime, Mip-NeRF~\cite{barron2021mipnerf} and its
% successors~\cite{barron2023zipnerf, muller2022instantngp} introduced continuously-valued scale and multi-resolution hash grids.  
% In the
% Gaussian regime, 
Mip Splatting~\cite{yu2024mipsplatting} constrains
primitive bandwidth for alias-free rendering, Hierarchical 3DGS~\cite{kerbl2024hierarchical} and Octree-GS~\cite{ren2024octreegs} construct explicit LOD hierarchies that select subsets by camera distance, LODGE~\cite{kulhanek2025lodge} and CLoD-GS~\cite{cheng2026clodgs} extend this to continuous distance-based filtering, Scaffold-GS~\cite{lu2024scaffoldgs} predicts view-adaptive
attributes from anchor features, and SMERF~\cite{duckworth2024smerf} demonstrates that hierarchical partitioning enables memory-bounded streaming on commodity devices. Each of these allocation criteria is geometric (global significance, camera distance, local redundancy, photometric error), and none consider what the robot's task requires.

\subsection{Active perception and non-uniform reconstruction}
Active view selection has been integrated with radiance fields: ActiveNeRF~\cite{pan2022activenerf} uses learned rendering uncertainty to rank candidate viewpoints, and FisherRF~\cite{jiang2024fisherrf} provides a Fisher Information criterion for next-best-view selection with a 3DGS backend. These methods decide \emph{where to look} but reconstruct uniformly within each selected view. VISTA~\cite{nagami2026vista} extends active perception to the trajectory level, steering the robot toward task-relevant objects specified by class query, though the underlying scene is still mapped uniformly. Foveated rendering adapts rendering effort to the observer's gaze: DeepFovea~\cite{kaplanyan2019deepfovea} reconstructs the periphery from sparse samples, VR-Splatting~\cite{franke2025vrsplatting} pairs low-primitive Gaussians in the periphery with neural points in the fovea for 90 Hz VR, and RTGS~\cite{lin2025rtgs} accelerates 3DGS-SLAM on edge devices through adaptive pruning and hardware-aware rendering optimizations. These systems require eye-tracking hardware to supply the gaze signal. GaussianLens~\cite{weng2025gaussianlens} performs on-demand densification of a pre-built low-resolution 3DGS model within a user-specified 2D region of interest, but operates offline after an initial uniform reconstruction. These methods vary where to look, how to render, or which region to refine, but none reallocate representation capacity during online mapping in response to a task.

\subsection{Language-conditioned 3D perception}
Vision-language models such as CLIP~\cite{radford2021clip} and class-agnostic segmentation networks such as SAM~\cite{kirillov2023sam} have enabled open-set 3D scene understanding.  LERF~\cite{kerr2023lerf} constructs a radiance field that renders dense CLIP feature maps queryable via natural-language text, and LangSplat~\cite{qin2024langsplat} extends this to Gaussian splatting with a substantial speedup.  At the object level, ConceptFusion~\cite{jatavallabhula2023conceptfusion} generates semantic 3D point clouds by assigning CLIP vectors to each point, and ConceptGraphs~\cite{gu2024conceptgraphs} builds 3D scene graphs with LLM-assigned relational edges. Semantic Gaussians~\cite{guo2024semanticgaussians} attaches CLIP vectors directly to each Gaussian primitive, and OpenGaussian~\cite{wu2024opengaussian} trains instance features with a codebook to cluster 3D Gaussians for open-set object extraction. 
Clio~\cite{maggio2024clio} formulates task-driven 3D scene understanding via the Information Bottleneck principle, clustering 3D primitives into task-relevant objects and regions in real time, and Bayesian Fields~\cite{maggio2025bayesianfields}  extends this line to 3DGS by fusing CLIP embeddings across views using Bayesian updating. These methods embed language directly into the scene representation; GaussLite instead uses a language-conditioned attention signal to drive the spatial allocation of Gaussian capacity during online mapping.

\section{Methodology}
\label{sec:method}

\begin{figure*}[t]
  \centering \includegraphics[width=0.95\linewidth, trim=1.1cm 2cm 1cm 2cm, clip]{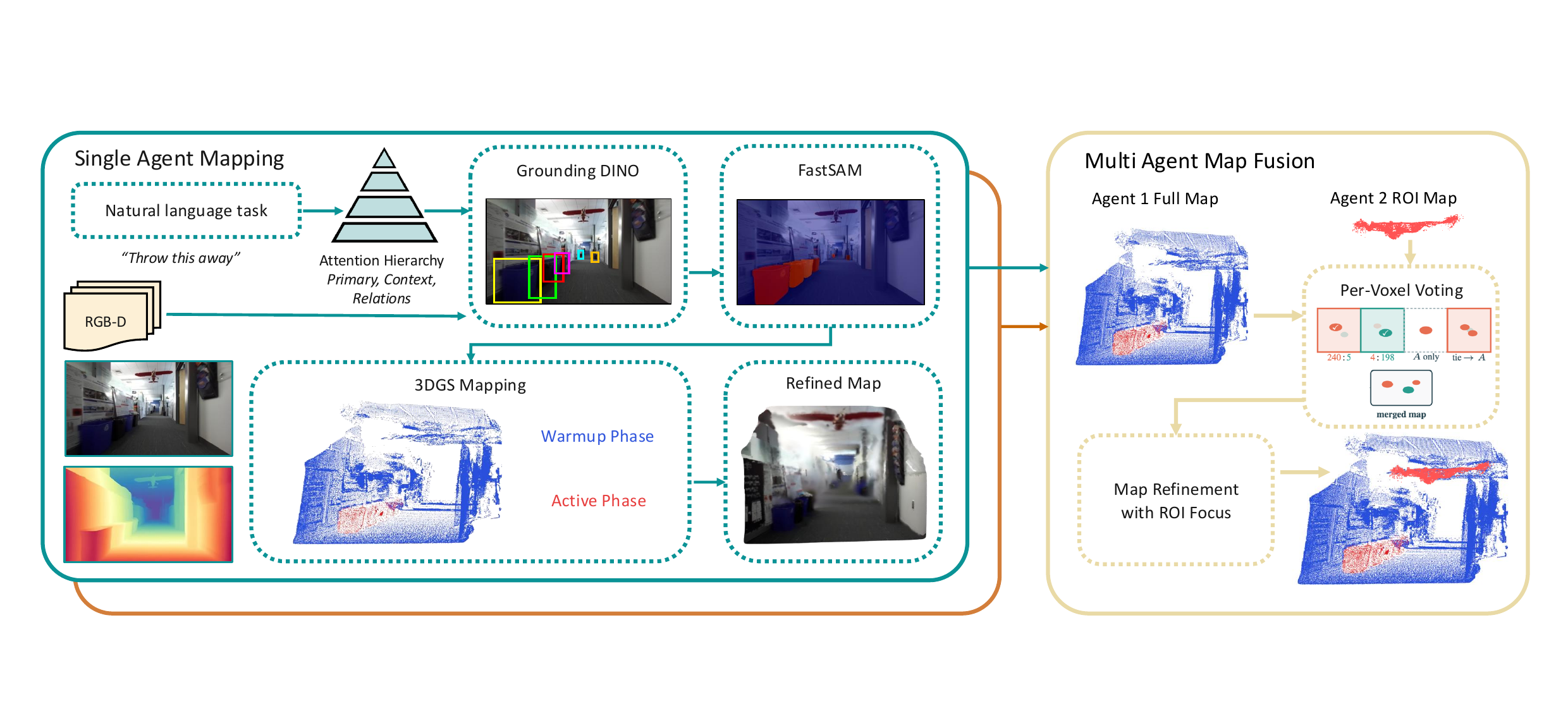}
  \caption{\textbf{System overview.}  A task description is parsed once into a
  structured object--relation graph.  Per-frame, an open-vocabulary detector
  and segmentor produce attention masks that modulate Gaussian seeding,
  initialization, and optimization within the mapping loop.}
  \vspace{-0.2cm}
  \label{fig:pipeline}
\end{figure*}

% system overview paragraph
\ours takes as input a sequential RGB-D stream with known camera poses and a natural-language task description, and incrementally builds a 3DGS map in which Gaussian density is concentrated on task-relevant geometry. \figref{fig:pipeline} gives an overview.  We first describe the Gaussian scene representation (\secref{sec:representation}) and the task-conditioned attention front-end that produces the per-pixel relevance map $w$ (\secref{sec:attention}). The remaining sections then introduce six methodological components, each isolated in the ablation of \tabref{tab:ablation}: ROI Seed Budgeting (\textbf{SB}), Stratified Initial Scaling (\textbf{SIS}), and Gradient-Focused Densification (\textbf{GFD}) govern \emph{where} new Gaussians are seeded (\secref{sec:mapping}); ROI Optimization (\textbf{ROI}) and Multi-View Optimization (\textbf{MVO}) govern \emph{how} they are trained (\secref{sec:optimization}); and multi-agent refinement merges task-specialized maps across robots (\secref{sec:fusion_method}).

% --------------------------------------------------------------------------
% 3.1 Gaussian Scene Representation
% --------------------------------------------------------------------------

\subsection{Gaussian Scene Representation}
\label{sec:representation}

Following~\cite{kerbl2023gaussiansplatting}, we represent the scene as a set of $N$ anisotropic 3D Gaussians $\mathcal{G} = \{G_i\}_{i=1}^{N}$, each parameterized by a mean position $\boldsymbol{\mu}_i \in \mathbb{R}^3$, a covariance $\boldsymbol{\Sigma}_i = \mathbf{R}_i \mathbf{S}_i \mathbf{S}_i^\top \mathbf{R}_i^\top$ factored into rotation $\mathbf{R}_i$ and diagonal scale $\mathbf{S}_i$, an opacity $\alpha_i \in [0,1]$, and an RGB color $\mathbf{c}_i \in [0,1]^3$. Rendering uses the differentiable rasterizer of~\cite{kerbl2023gaussiansplatting}, which produces a color image $\hat{I}$, depth $\hat{D}$, and alpha-accumulation map $\hat{\alpha}$.
Beyond these standard 3DGS attributes, each $G_i$ carries two non-differentiable task-related variables: an active flag $a_i \in \{0,1\}$ set at seed time from the binary relevance map, which controls whether $G_i$ receives gradient during the active optimization phase; and an active-optimization counter $\kappa_i \in \mathbb{N}$, which is incremented each iteration that $G_i$ is in the active set, which serves as the fusion weight for multi-agent merging (\secref{sec:fusion_method}).

% --------------------------------------------------------------------------
% 3.2 Task-Conditioned Attention
% --------------------------------------------------------------------------

\subsection{Task-Conditioned Attention Front-End}
\label{sec:attention}

\subsubsection{Structured task parsing}
Given a free-form task description $\mathcal{T}$ (\eg, ``\textit{prepare to pick up the object on the desk}''), we prompt a lightweight instruction-tuned LLM (Phi-3-mini \cite{microsoft2024phi3}, 3.8\,B parameters) with a few-shot schema to extract a
structured task graph $\mathcal{G}_\text{task} = \{\mathcal{O}_\text{target}, \mathcal{O}_\text{anchor}, \mathcal{R}\}$, where $\mathcal{O}_\text{target}$ are the target noun phrases (\eg, \emph{object}), $\mathcal{O}_\text{anchor}$ are the spatial reference objects against which targets are constrained (\eg, \emph{desk}), and $\mathcal{R}$ contains binary spatial predicates of the form $r(t, a)$ with $t \in \mathcal{O}_\text{target}$, $a \in \mathcal{O}_\text{anchor}$; we implement two predicates, \textsc{near} (Euclidean distance below a threshold) and \textsc{on} (target centroid above anchor centroid with overlapping $xy$ projection), which span the spatial constraints in our evaluation tasks. We restrict $\mathcal{R}$ to centroid-based predicates because they remain well-defined under partial observation; richer predicates (\textsc{inside}, \textsc{behind}) require fuller object geometry and are deferred to future work. Parsing runs once per task ($<$4\,s) and produces the query set $\mathcal{Q} = \mathcal{O}_\text{target} \cup \mathcal{O}_\text{anchor}$.

\subsubsection{Open-vocabulary grounding and segmentation}
% On each motion-triggered keyframe (translation $> \theta_t{=}1.0$\,m or rotation $> \theta_r{=}15^{\circ}$), we run Grounding DINO~\cite{liu2024groundingdino} ($\texttt{box\_threshold}{=}0.20$, $\texttt{text\_threshold}{=}0.25$) with the query set $\mathcal{Q}$ to produce 2D bounding boxes.  Boxes exceeding a \texttt{max\_box\_fraction}
% of image area (0.20 for small objects, 0.40 for furniture, 0.70 for multi-object tasks) are discarded.  Surviving boxes are refined into pixel-level masks by FastSAM ($\texttt{conf}{=}0.4$,
% $\texttt{iou}{=}0.9$), matched to boxes by IoU.  Between detection frames, prior boxes are reprojected via their 3D frustum corners and re-segmented by FastSAM, avoiding the full Grounding DINO forward pass.  The text encoder is cached once per mission.
On each motion-triggered keyframe, we run Grounding DINO~\cite{liu2024groundingdino} with the query set $\mathcal{Q}$ to produce 2D bounding boxes, filtering out detections that span too much of the image. Resulting boxes are refined into pixel-level masks by FastSAM~\cite{zhao2023fast}, matched to boxes by IoU. Between detection frames, prior boxes are re-projected via their 3D frustum corners and re-segmented by FastSAM, avoiding the full Grounding DINO forward pass.

\begin{figure}[t]
\centering
\includegraphics[width=1\linewidth, trim = {0.5cm 0 0.5cm 0}, clip]{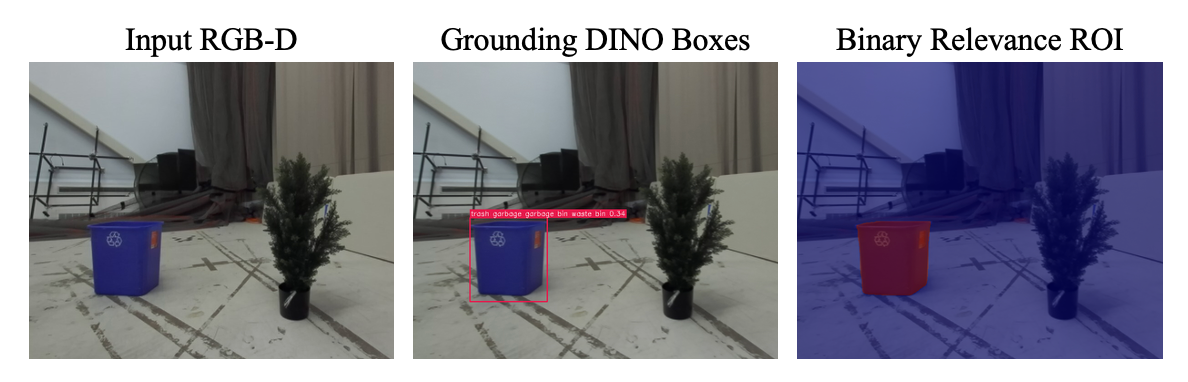}
\caption{Task-to-attention front-end on Replica . Grounding DINO detects task-relevant objects; FastSAM produces pixel-level
masks; spatial filtering prunes detections inconsistent with the 3D map.}
\vspace{-0.5cm}
\label{fig:attention}
\end{figure}

\subsubsection{3D spatial filtering}
Spatial predicates in $\mathcal{R}$ are relational and require 3D positions for both target and anchor objects. For each detection $i$, we obtain a 3D centroid $\mathbf{c}_i$ by lifting the 2D mask centroid into world coordinates through the depth rendered from the current Gaussian map. The predicate is then evaluated against the anchor centroids:
$$\text{keep}(i) = \bigvee_{r \in \mathcal{R}} \text{pred}_r\bigl(\mathbf{c}_i,; {\mathbf{c}_j : j \in \text{anchors}(r)}\bigr)$$
with predicates \textsc{near}($d < \tau$) and
\textsc{on}($\Delta z > 0 \wedge d_{xy} < \epsilon$).

\subsubsection{Per-pixel relevance map}
Resulting masks are merged into a binary per-pixel relevance map
$w(u,v) \in \{0, 1\}$.  Each new Gaussian inherits its active flag $a_i$ from $w$ at its seed pixel: active Gaussians ($a_i{=}1$) receive concentrated optimization, while background Gaussians ($a_i{=}0$) serve as a coarse scaffold.  \figref{fig:attention} illustrates the pipeline.

% --------------------------------------------------------------------------
% 3.3 Relevance-Driven Seeding (D, SIS, GFD)
% --------------------------------------------------------------------------
% (where Gaussians go) (1) ROI Densification (D), (2) Stratified Initial Scaling (SIS), (3) Gradient-Focused Densification (GFD).
\subsection{Relevance-Driven Seeding}
\label{sec:mapping}

We build on the Gaussian-SLAM~\cite{yugay2024gaussianslam} mapper and treat camera poses as an external input, decoupling our mapping contribution from pose estimation; on Replica we use simulation ground-truth poses, while on our campus dataset we use LiDAR-inertial odometry from DLIO~\cite{chen2023dlio}. We maintain a single global submap so that task-relevant Gaussians co-train against every keyframe that observes them. Each frame contributes $N$ new Gaussians by back-projecting sampled depth pixels into world coordinates. Three components govern the seed budget and its spatial layout:

\subsubsection{ROI Seed Budgeting (SB)}
\label{sec:densification_D}
The seed budget is split between ROI and background pixels by an ROI allocation $\rho \in [0,1]$: $N_\text{ROI} = \lfloor \rho N \rfloor$ seeds are drawn from pixels with $w(u,v) = 1$ and the remaining $N_\text{bg} = N - N_\text{ROI}$ from background. This concentrates Gaussians on task-relevant geometry while retaining a coarse background scaffold for geometric anchoring and free-space coverage.
% ($\rho{=}40$ on Replica)

\subsubsection{Stratified Initial Scaling (SIS)}
\label{sec:sis}
Two coupled details govern how the partitioned seeds are realized in space and scale. First, background pixels are drawn via stratified sampling: the non-ROI image area is partitioned into a $\sqrt{N_\text{bg}} \times \sqrt{N_\text{bg}}$ grid with one jittered sample per cell, eliminating the clumping artifacts of uniform random sampling at fixed seed budget. Second, each new Gaussian is initialized with identity rotation, color from the observed RGB, opacity $\alpha_0 = 0.5$, and an initial scale proportional to the local median nearest-neighbor distance $s_0$, modulated by a per-tier multiplier:
\vspace{-0.2cm}
\begin{equation}
  \mathbf{s}_\text{init} = \log\bigl(s_0 \cdot m_\tau\bigr) \cdot \mathbf{1},
  \quad
  m_\tau =
  \begin{cases}
    0.5 & \text{ROI seed} \\
    0.7 & \text{background seed}.
  \end{cases}
  % \label{eq:scaleinit}
\end{equation}
The smaller ROI scale allows finer detail; the reduced background scale prevents alpha-bleed into neighboring ROI pixels. Under a limited per-frame iteration budget, the optimizer cannot shrink oversized Gaussians, so starting small is critical (\secref{sec:ablations}).

\subsubsection{Gradient-Focused Densification (GFD)}
\label{sec:gfd}
Within the seed budget assigned by \textbf{SB} and the cells imposed by \textbf{SIS}, pixel selection within each cell is biased toward high-frequency image content. We compute a per-keyframe edge-magnitude map $E(u,v) = \|\nabla I(u,v)\|$ via a Sobel filter on the luminance channel and turn it into a within-cell sampling distribution, where within each stratified cell, the seed pixel is sampled with probability proportional to $E(u,v) + \epsilon$, where $\epsilon$ is a small smoothing constant. The same gradient weighting is applied when sampling the ROI seed pool. The effect is to place Gaussians at object silhouettes and textured surfaces, where photometric residual is highest and where additional density yields the largest increase in rendering fidelity (\tabref{tab:ablation}).

% --------------------------------------------------------------------------
% 3.4 Relevance-Driven Optimization (ROI, MVO)
% --------------------------------------------------------------------------
% (how they are trained): (1) Two-Phase ROI Optimization (ROI), (2) Multi-View Optimization (MVO), (3) Loss and optimizer, (4) Adaptive density control.

\subsection{Relevance-Driven Optimization}
\label{sec:optimization}

\subsubsection{ROI Optimization (ROI)}
\label{sec:roi_opt}
For each mapped frame we run a brief warmup optimization over the full image, then restrict both rendering and gradient updates to active Gaussians ($a_i = 1$) for the remaining iterations. Excluding background Gaussians from the active phase concentrates optimization budget on task-relevant geometry and roughly halves the per-iteration render cost, since the rasterizer processes only the ROI subset. Each active Gaussian's counter $\kappa_i$ is incremented per active-phase iteration. The per-iteration loss combines absolute-error color and depth terms with an SSIM term:
\begin{equation}
  \mathcal{L} = \mathcal{L}_\text{color} + \lambda_d \, \mathcal{L}_\text{depth}
    + \lambda_s \, (1 - \text{SSIM}(\hat{I}, I)).
  \label{eq:loss}
\end{equation}
The color and SSIM terms are evaluated over the full image during warmup and restricted to ROI pixels intersected with the rendering footprint during the active phase.

\begin{table*}[t]
  \centering
  \caption{\textbf{ROI Rendering Performance on Replica:} External baselines are
  budget-matched at maximum 1M Gaussians and $\leq0.5$\,s/frame. Results show per-scene averages across manipulation, navigation and object search tasks. }  
  \label{tab:main_quality}
  \footnotesize
  \setlength{\tabcolsep}{11pt}
  \renewcommand{\arraystretch}{1.0}

  \begin{tabular}{c l c c c c c c c c c c}
    \toprule
    \textbf{Method} & \textbf{Metric}
      & \textbf{Rm0} & \textbf{Rm1} & \textbf{Rm2}
      & \textbf{Off0} & \textbf{Off1} & \textbf{Off2}
      & \textbf{Off3} & \textbf{Off4} & \textbf{Avg.} \\
    \midrule

    % \multirow{12}{*}{\rotatebox[origin=c]{90}{\textbf{Navigation}}}

    \multirow{3}{*}{Gaussian-SLAM~\cite{yugay2024gaussianslam}}
      & PSNR$\uparrow$
      & 24.37 & \underline{25.38} & 28.05 & 31.31 & 30.77 & 23.99 & 23.74 & 25.39 & 26.62 \\
      & SSIM$\uparrow$
      & 0.767 & \underline{0.738} & \underline{0.857} & 0.877 & \textbf{0.898} & 0.825 & 0.788 & 0.836 & \underline{0.823} \\
      & LPIPS$\downarrow$
      & 0.355 & 0.380 & 0.306 & 0.268 & \underline{0.187} & 0.308 & 0.326 & 0.298 & 0.303 \\

    \noalign{\vskip 2pt}
    \cdashline{1-11}
    \noalign{\vskip 2pt}

    \multirow{3}{*}{SplaTAM~\cite{keetha2024splatam}}
      & PSNR$\uparrow$
      & \underline{25.58} & 23.41 & \underline{29.19} & \underline{35.57} & \underline{31.36} & 21.16 & \textbf{26.32} & \underline{29.91} & \underline{28.56} \\
      & SSIM$\uparrow$
      & \textbf{0.801} & 0.715 & 0.821 & \underline{0.898} & 0.863 & \textbf{0.850} & 0.804 & 0.818  & 0.821 \\
      & LPIPS$\downarrow$
      & \textbf{0.247} & \underline{0.311} & \underline{0.254} & \textbf{0.160} & 0.191 & \textbf{0.204} & \textbf{0.224} & \underline{0.247}  & \textbf{0.230} \\

    \noalign{\vskip 2pt}
    \cdashline{1-11}
    \noalign{\vskip 2pt}

    \multirow{3}{*}{MonoGS~\cite{matsuki2024monogs}}
      & PSNR$\uparrow$
      & 24.24 & 24.57 & 27.14 & 31.57 & 27.28 & 23.26 & 24.87 & 25.83 & 26.10 \\
      & SSIM$\uparrow$
      & 0.777 & 0.731 & 0.848 & \textbf{0.912} & 0.784 & 0.817 & \underline{0.809} & \underline{0.846} & 0.815 \\
      & LPIPS$\downarrow$
      & 0.322 & 0.396 & 0.303 & 0.191 & 0.233 & 0.275 & 0.286 & 0.286 & 0.287 \\

    \noalign{\vskip 2pt}
    \cdashline{1-11}
    \noalign{\vskip 2pt}

    % ROI WEIGHTED RESULTS: 
    % \multirow{3}{*}{\ours}
    %   & PSNR$\uparrow$
    %   & \textbf{26.47} & \textbf{26.47} & \textbf{31.46} & \textbf{37.22} & \textbf{31.44} & \textbf{28.59} & \underline{26.14} & \textbf{30.72} & \textbf{29.81} \\
    %   & SSIM$\uparrow$
    %   & \underline{0.773} & \textbf{0.775} & \textbf{0.860} & 0.887 & \underline{0.891} & \underline{0.827} & \textbf{0.812} & \textbf{0.853} & \textbf{0.835} \\
    %   & LPIPS$\downarrow$
    %   & \underline{0.304} & \textbf{0.283} & \textbf{0.239} & \underline{0.183} & \textbf{0.135} & \underline{0.247} & \underline{0.235} & \textbf{0.214} & \textbf{0.230} \\

    \multirow{3}{*}{\ours}
      & PSNR$\uparrow$
      & \textbf{26.47} & \textbf{26.47} & \textbf{31.46} & \textbf{37.22} & \textbf{31.44} & \textbf{28.59} & \underline{26.14} & \textbf{30.72} & \textbf{29.81} \\
      & SSIM$\uparrow$
      & \underline{0.773} & \textbf{0.775} & \textbf{0.860} & 0.887 & \underline{0.891} & \underline{0.827} & \textbf{0.812} & \textbf{0.853} & \textbf{0.835} \\
      & LPIPS$\downarrow$
      & \underline{0.304} & \textbf{0.283} & \textbf{0.239} & \underline{0.183} & \textbf{0.135} & \underline{0.247} & \underline{0.235} & \textbf{0.214} & \textbf{0.230} \\

    \bottomrule
  \end{tabular}
  \vspace{-0.3cm}
\end{table*}

\subsubsection{Multi-View Optimization (MVO)}
\label{sec:mvo}
At each iteration, the training view is sampled uniformly at random from a window of recent keyframes. This prevents over-supervision by the current viewpoint and forces ROI Gaussians to remain consistent across recent observations~\cite{keetha2024splatam}. We optimize with Adam using per-parameter-group learning rates and apply the standard 3DGS clone/split/prune scheme~\cite{kerbl2023gaussiansplatting}, with cloned and split daughters inheriting the parent's $a_i$.
\subsection{Multi-Agent Refinement}
\label{sec:fusion_method}

% \begin{figure}
%     \centering
%     \includegraphics[width=0.99\linewidth, trim={0.5cm 0 0 0}, clip]{Figures/fusion_v2.png}
%     \caption{\textbf{Multi-Agent Refinement.} Agent $2$ shares only its ROI Gaussians with Agent $1$. At each $\delta{=}0.10$\,m voxel, the source with the higher aggregate $\kappa_i$ is kept, so well-optimized ROI Gaussians win over the other agent's low-count background. In this example, Agent 1 maps boxes and Agent 2 shares high-fidelity trash cans which replace Agent 1's voxels after fusion. \todo{use actual map renders for fuzzy background}}
%     \label{fig:voxel_vote}
% \end{figure}

% \begin{figure*}[t]
%     \centering
%     \includegraphics[width=0.95\linewidth]{Figures/dataset_preview.png}
%     \caption{Caption}
%     \label{fig:placeholder}
% \end{figure*}

When multiple agents map the same environment with different task specifications, their Gaussian maps can be fused without re-optimization. The counter $\kappa_i$ serves as a per-Gaussian quality proxy: ROI Gaussians accumulate high $\kappa_i$ while background Gaussians remain near zero. Multi-agent refinement consists of a count-weighted voxel vote that produces an initial merged map, followed by an optional brief joint optimization pass that smooths boundaries between the agents' contributions.

We discretize world space into a voxel grid and, at each voxel where both agents contributed Gaussians, retain the source whose aggregate $\kappa_i$ is higher; non-overlapping voxels carry over unchanged, and ties go to the denser source. The intuition is that ROI Gaussians accumulate high counts during their agent's active phase while background Gaussians remain near zero, so the vote picks the better-trained representation per region. Because only the ROI subset carries meaningful counts, the effective data exchange is a small fraction of the full map.

The voxel vote operates at the resolution $\delta$ of the voxel grid and can leave seams where adjacent voxels were won by different agents. After voting, we can run a joint optimization pass over a keyframe pool drawn uniformly from the union of agents' keyframe buffers, using the same loss (Eq.~\eqref{eq:loss}) and adaptive density control as the per-frame mapper. During this pass, all resulting Gaussians are treated as active so that both agents' contributions are jointly refined at shared regions. The refinement is optional: the voxel vote alone yields a usable merged map for online operation, while the refinement pass is invoked when the team has spare compute and additional ROI fidelity is desired (\tabref{tab:fusion}).

\section{Evaluation}
\label{sec:experiments}

% --------------------------------------------------------------------------
% 4.1 Setup
% --------------------------------------------------------------------------
\subsection{Experimental Setup}
\label{sec:setup}

\begin{table*}[t]
  \centering
  \caption{\textbf{ROI Rendering Performance on Campus Dataset:} External baselines are
  budget-matched at maximum 3M Gaussians and real time mapping. Single-Agent tasks show average of single-agent performance. Multi-Agent tasks are evaluated on fused ROIs. }  
  % \label{tab:main_quality}
  \footnotesize
  \setlength{\tabcolsep}{14pt}
  \renewcommand{\arraystretch}{1.0}

  \begin{tabular}{c c l c c c c c c c c c c c}
    \toprule
    \textbf{Method} & \textbf{Task} & \textbf{Metric}
      & \textbf{Hb1} & \textbf{Hb2} & \textbf{Off} & \textbf{Od1} & \textbf{Od2} & \textbf{Avg.} \\
    \midrule

    % \multirow{12}{*}{\rotatebox[origin=c]{90}{\textbf{Navigation}}}

    \multirow{3}{*}{Gaussian-SLAM~\cite{yugay2024gaussianslam}}
      & \textbf{Single-Agent} & PSNR$\uparrow$
      & 18.78 & 19.28 & 20.83 & 15.65 & 15.80 & 18.06 \\
      & & SSIM$\uparrow$
      & 0.533 & 0.565 & 0.682 & 0.404 & 0.392 & 0.515 \\
      & & LPIPS$\downarrow$
      & 0.534 & 0.509 & 0.419 & 0.743 & 0.750 & 0.591 \\

    \noalign{\vskip 2pt}
    \cdashline{1-11}
    \noalign{\vskip 2pt}

    \multirow{3}{*}{SplaTAM~\cite{keetha2024splatam}}
      & & PSNR$\uparrow$
      & 20.16 & 20.65 & \underline{22.96} & \underline{16.67} & \underline{16.34} & \underline{19.35}  \\
      & & SSIM$\uparrow$
      & \underline{0.596} & \textbf{0.644} & 0.714 & \textbf{0.523} & \textbf{0.555} & \textbf{0.606}  \\
      & & LPIPS$\downarrow$
      & 0.546 & 0.515 & \underline{0.334} & \textbf{0.509} & \textbf{0.503} & \underline{0.481}   \\

    \noalign{\vskip 2pt}
    \cdashline{1-11}
    \noalign{\vskip 2pt}

    \multirow{3}{*}{MonoGS~\cite{matsuki2024monogs}}
      & & PSNR$\uparrow$
      & \underline{21.21} & \underline{21.77} & 22.02 & 15.53 & 14.81 & 19.06  \\
      & & SSIM$\uparrow$
      & \textbf{0.632} & \underline{0.630} & \textbf{0.742} & 0.380 & 0.369 & 0.550 \\
      & & LPIPS$\downarrow$
      & \textbf{0.493} & \underline{0.481} & 0.394 & 0.736 & 0.753 & 0.571  \\

    \noalign{\vskip 2pt}
    \cdashline{1-11}
    \noalign{\vskip 2pt}

    \multirow{3}{*}{\ours}
      & & PSNR$\uparrow$
      & \textbf{21.95} & \textbf{23.91} & \textbf{24.63} & \textbf{17.44} & \textbf{17.31} & \textbf{21.05}  \\
      & & SSIM$\uparrow$
      & 0.582 & 0.609 & \underline{0.725} & \underline{0.487} & \underline{0.496} & \underline{0.580} \\
      & & LPIPS$\downarrow$
      & \underline{0.502} & \textbf{0.473} & \textbf{0.322} & \underline{0.546} & \underline{0.514} & \textbf{0.471}  \\

    \midrule

    \multirow{3}{*}{Gaussian-SLAM~\cite{yugay2024gaussianslam}}
      & \textbf{Multi-Agent} & PSNR$\uparrow$
      & 19.04 & 19.66 & - & 15.11 & 15.83 & 17.41   \\
      & & SSIM$\uparrow$
      & 0.566 & 0.597 & - & 0.386 & 0.386 & 0.484 \\
      & & LPIPS$\downarrow$
      & 0.549 & 0.504 & - & 0.793 & 0.784 & 0.658 \\

    \noalign{\vskip 2pt}
    \cdashline{1-11}
    \noalign{\vskip 2pt}

    \multirow{3}{*}{SplaTAM~\cite{keetha2024splatam}}
      & & PSNR$\uparrow$
      & 19.98 & 20.44 & - & \underline{16.22} & \underline{16.10} & 18.19  \\
      & & SSIM$\uparrow$
      & 0.602 & \underline{0.652} & - & \underline{0.504} & \underline{0.542} & \underline{0.575} \\
      & & LPIPS$\downarrow$
      & 0.533 & 0.494 & - & \underline{0.549} & \underline{0.543} & \underline{0.530}  \\

    \noalign{\vskip 2pt}
    \cdashline{1-11}
    \noalign{\vskip 2pt}

    \multirow{3}{*}{MonoGS~\cite{matsuki2024monogs}}
      & & PSNR$\uparrow$
      & \underline{21.43} & \underline{22.40} & - & 14.76 & 14.49 & \underline{18.27} \\
      & & SSIM$\uparrow$
      & \underline{0.645} & 0.645 & - & 0.350 & 0.353 & 0.498 \\
      & & LPIPS$\downarrow$
      & \underline{0.510} & \underline{0.478} & - & 0.772 & 0.778 & 0.635 \\
    \noalign{\vskip 2pt}
    \cdashline{1-11}
    \noalign{\vskip 2pt}

    \multirow{3}{*}{\ours}
      & & PSNR$\uparrow$
      & \textbf{22.74} & \textbf{22.95} & - & \textbf{19.89} & \textbf{19.95} & \textbf{21.38} \\
      & & SSIM$\uparrow$
      & \textbf{0.727} & \textbf{0.709} & - & \textbf{0.690} & \textbf{0.699} & \textbf{0.706} \\
      & & LPIPS$\downarrow$
      & \textbf{0.432} & \textbf{0.455} & - & \textbf{0.450} & \textbf{0.454} & \textbf{0.448}  \\

    \bottomrule
  \end{tabular}
  \vspace{-0.4cm}
  \label{tab:campus-quality}
\end{table*}

\subsubsection{Datasets}
We evaluate \ours against existing Gaussian SLAM approaches on the Replica dataset and on an in-house dataset we call Campus. Replica~\cite{straub2019replica} provides eight photorealistic indoor RGB-D sequences with clean depth and color; we use the simulator's ground-truth poses across all baselines to isolate mapping quality from tracking error. Campus is comprised of three scenes including a highbay, an office, and an outdoor setting (Fig.~\ref{fig:pipeline}; Fig.~\ref{fig:qualitative}), captured onboard a Clearpath Husky rover with a ZED 2i stereo camera and processed on an RTX 4070 Laptop GPU. Pose estimates are produced online by DLIO~\cite{chen2023dlio}, demonstrating performance from real onboard sensing.

\subsubsection{Task specifications}
We evaluate \ours across navigation, manipulation, and semantic-search tasks. On Replica we use \textit{``Prepare to pick up the object on the desk''} (manipulation), \textit{``Identify the chair closest to the table''} (search), and \textit{``Locate all obstacles in the scene and find a path to the door''} (navigation), yielding 24 scene–task evaluation pairs. On Campus we use \textit{``Pick up the box''} (manipulation), \textit{``Map the lawn chairs''} (search), and \textit{``Avoid the trash cans/dirt''} (navigation), yielding 9 scene–task evaluation pairs.

% \begin{itemize}
%   \item \textbf{Navigation:}  ``Go to the [landmark].''  Primary attention on
%   the navigation target (door, desk, window); secondary on obstacles along the
%   path.
%   \item \textbf{Manipulation:}  ``Pick up the [object] from the [surface].''
%   Primary attention on the grasp target; secondary on the supporting surface;
%   context on surrounding collision geometry.
%   \item \textbf{Semantic search:}  ``Find the [object].''  Attention is
%   initially diffuse and sharpens as the object enters the field of view.
% \end{itemize}

\subsubsection{Baselines}
We compare against three external baselines: Gaussian-SLAM~\cite{yugay2024gaussianslam}, SplaTAM~\cite{keetha2024splatam}, and MonoGS~\cite{matsuki2024monogs}. Together they represent state-of-the-art Gaussian Splatting systems: Gaussian-SLAM is the mapper \ours{} is built on, so the comparison isolates our task-conditioning contribution; SplaTAM is a canonical RGB-D 3DGS-SLAM; and MonoGS pushes a different design lever (aggressively small initial Gaussian scale). All three have publicly available implementations.

Each baseline is run from its released configuration, with the per-frame iteration count and frame-step interval adjusted to meet the same real-time per-frame budget as \ours. All comparisons use the same pose source, a matched Gaussian budget, and no offline refinement.
%   backbone with uniform seeding, tuned to matched budget via
%   \texttt{new\_submap\_every}$=$500, 4 submaps, 7 iterations/frame.
%   \item \textbf{SplaTAM}~\cite{keetha2024splatam}: online dense RGB-D SLAM
%   with per-keyframe Gaussian addition, budget-matched via initial stride$=$2
%   and per-keyframe add cap of 2000.
%   \item \textbf{MonoGS}~\cite{matsuki2024monogs}: monocular Gaussian SLAM
%   with continuous densification and small initial Gaussian scale
%   (\texttt{point\_size}$=$0.05), with offline color refinement disabled.
% \end{itemize}
% We also include an \textbf{internal uniform baseline}: our own codebase
% with uniform seeding (no task-conditioned allocation), multi-submap
% architecture, and 8 iterations/frame.  This baseline uses ${\sim}140$k
% Gaussians at its default configuration and is \emph{not} budget-matched; it
% is included to show the performance of the codebase without any
% task-conditioned components.

\subsubsection{Metrics}
We report rendering quality \emph{stratified by task relevance} using hand-annotated ROI masks including PSNR, SSIM, LPIPS computed within ground-truth task-relevant masks only. We note that each of these methods are capped in terms of map size to 1M Gaussians on scenes in the Replica dataset and to 3M on scenes in the Campus dataset as well as to real-time performance on an NVIDIA GeFORCE RTX 4070 Laptop GPU. 

% --------------------------------------------------------------------------
% 4.2 Main Results
% --------------------------------------------------------------------------
\subsection{Main Results}
\label{sec:results}
\vspace{-0.05cm}

\begin{figure*}[t]
\centering
\includegraphics[width=\linewidth, trim = {3cm 15.5cm 2.5cm 0}, clip]{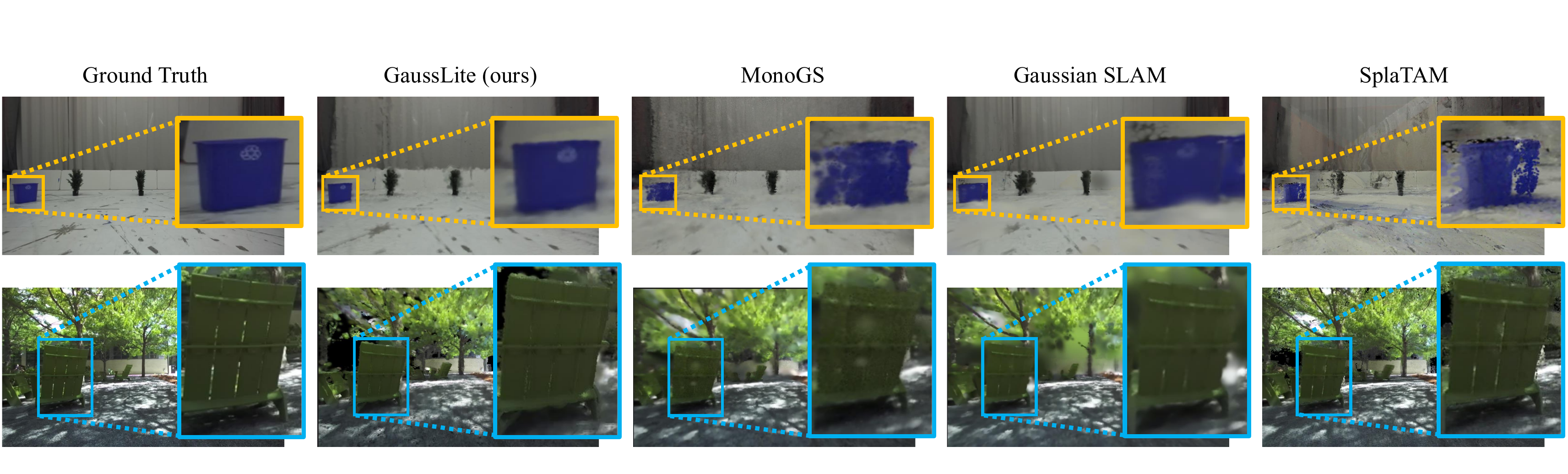}
% \placeholderfig[5.5cm]{%
% 2 rows (scenes) $\times$ 5 columns (G-SLAM, SplaTAM, MonoGS, GaussLite, GT). crop insets on ROI. Rows: From campus data to show dataset}
\caption{Qualitative comparison on Campus Dataset.  Insets show crops of task-relevant regions.  \ours preserves fine detail in task-relevant regions (comparable to ground truth) while baselines either lose foreground detail after uniform allocation, or produce sharper edges but lower pixel fidelity in the ROI. }
\label{fig:qualitative}
\vspace{-0.2cm}
\end{figure*}

On Replica (\tabref{tab:main_quality}), among the three budget-matched external baselines, \ours{} achieves the best mean ROI PSNR (29.81 dB) and SSIM (0.835), and ties SplaTAM for best LPIPS (0.230). \ours{} surpasses in PSNR on 7/8 scenes and outperforms SplaTAM by $+1.25$ dB, Gaussian-SLAM by $+3.19$ dB, and MonoGS by $+3.71$ dB in mean ROI PSNR. On the Campus dataset (Tab. \ref{tab:campus-quality}), \ours{} achieves the best ROI PSNR in all settings, outperforming SplaTAM by $+1.70$ dB, Gaussian-SLAM by $+2.99$ dB, and MonoGS by $+1.99$ dB in mean ROI PSNR. Because all comparisons are at matched Gaussian count and real-time per-frame budget, these gains are entirely attributable to where the budget is spent. For reference, $+1$ dB increase in ROI PSNR corresponds to approximately a $21\%$ reduction in mean-squared error (MSE) within task-relevant regions. Our observed improvements range from this level up to approximately a $74\%$ MSE reduction in some settings. Qualitative results are shown in Fig~\ref{fig:qualitative}.

Each method makes a different tradeoff between pixel-level accuracy (PSNR) and perceptual quality (SSIM, LPIPS), which are amplified under constrained budgets. For task-driven robotics, ROI PSNR is the most directly actionable metric, since downstream perception algorithms depend on the fidelity of the underlying pixel intensities. \ours{} wins on this primary metric in nearly every setting while remaining best-or-tied on the perceptual ones.

% --------------------------------------------------------------------------
% 4.3 Ablation Studies
% --------------------------------------------------------------------------
\subsection{Ablation Studies}
\label{sec:ablations}

\begin{figure}
    \centering
    \includegraphics[width=0.98\linewidth]{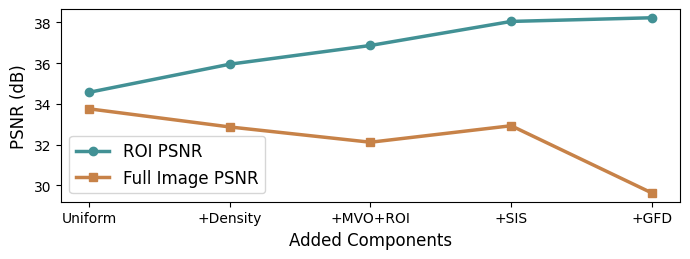}
    \caption{Ablation over tradeoff in full-image quality on each added component of \ours{} (SB, MVO, ROI, SIS, GFD) on Office0 from Replica.}
    \label{fig:ablation-fig}
    \vspace{-0.3cm}
\end{figure}

We ablate each component of the \ours{} pipeline by progressively enabling SB, MVO, ROI, SIS, and GFD on Replica Office0 and Campus Highbay (Table~\ref{tab:ablation}). To isolate the contribution of each budget-allocation component from the upstream task-attention pipeline, we run this ablation with ground-truth ROI masks, so the deltas reflect the mapper's response to a clean relevance signal rather than detection error. Every component contributes a positive ROI PSNR gain. The largest single jumps come from ROI Seed Budgeting ($+1.38$ dB on Replica) and Stratified Initial Scaling ($+1.18$ dB on Replica, $+2.99$ dB on Campus), with the Campus result showing that small-init Gaussians become especially important under real-world depth and exposure noise. GFD adds a smaller but consistent improvement on both datasets. The fully ablated pipeline lifts ROI PSNR from $34.57$ to $38.23$ dB on Replica and $21.50$ to $25.25$ dB on Campus. Fig.~\ref{fig:ablation-fig} shows the corresponding tradeoff in full-image rendering quality as each component is enabled. ROI PSNR rises monotonically, while full-image PSNR drops modestly as budget is reallocated away from the background, which is the intended tradeoff of task-conditioned mapping.

\subsection{Multi-Agent Fusion}
\label{sec:fusion}

% \begin{table}[t!]
% \centering
% \caption{Percentage of each agent's map that is task-relevant (ROI), shared between agents during multi-agent fusion \todo{check if we should include table or just inline text}}
% \label{tab:roi-fraction}
% \begin{tabular}{@{}llr@{}}
% \toprule
% \textbf{Group} & \textbf{Run} & \textbf{Map Shared \%}\\
% \midrule
% \multirow{4}{*}{Highbay} & Agent 1 (boxes)       &  5.13\% \\
%                          & Agent 1 (trash cans)  &  8.36\% \\
%                          & Agent 2 (boxes)       &  4.93\% \\
%                          & Agent 2 (trash cans)  & 29.14\% \\
% \midrule
% \multirow{4}{*}{Outdoor} & Agent 1 (lawn chairs) &  3.42\% \\
%                          & Agent 1 (avoid dirt)  &  1.44\% \\
%                          & Agent 3 (lawn chairs) &  1.09\% \\
%                          & Agent 3 (avoid dirt)  &  3.16\% \\
% \bottomrule
% \end{tabular}
% \vspace{-0.3cm}
% \end{table}

We evaluate the fusion algorithm of Sec.~\ref{sec:fusion_method} on two real-world sequences (Campus Highbay and Campus Outdoor) where two agents build task-specialized maps that must be merged into a single representation. We compare three strategies: \emph{Combine Maps} (naive concatenation), \emph{Voxel Voting} ($\kappa_i$-weighted per-voxel selection), and \emph{Refinement} (Voxel Voting followed by a short joint optimization pass). On the highbay sequence, our voxel vote alone yields near-specialist quality, and the refinement pass pushes this to $+1.74$ dB on Replica and $+4.16$ on Campus (\tabref{tab:fusion}). Because only the ROI subset carries meaningful $\kappa_i$ values, the inter-agent data exchange is on average only 7.08\% of the full map.

\begin{table}[t!]
\centering
\caption{\blue{Component-wise ablation of \ours{} on Office0 from Replica and Agent~1 of Campus Highbay, progressively enabling ROI Seed Budgeting (SB), Multi-View Optimization (MVO), ROI Optimization (ROI), Stratified Initial Scaling (SIS), and Gradient-Focused Densification (GFD) (ROI PSNR $\uparrow$ [dB]).}}
\footnotesize
\setlength{\tabcolsep}{8.5pt}
\begin{tabular}{cccccccc@{}}
\toprule
\textbf{SB} & \textbf{MVO} & \textbf{ROI} & \textbf{SIS} & \textbf{GFD} & \textbf{Replica} & \textbf{Campus}\\ \midrule
\redx        & \redx        & \redx        & \redx        & \redx         & 34.57          & 21.50 \\
\greencheck  & \redx        & \redx        & \redx        & \redx         & 35.95          & 21.61 \\
\greencheck  & \greencheck  & \redx        & \redx        & \redx         & 35.96          & 22.55 \\
\greencheck  & \greencheck  & \greencheck  & \redx        & \redx         & 36.87          & 22.15 \\
\greencheck  & \greencheck  & \greencheck  & \greencheck  & \redx         & 38.05          & 25.14 \\
\greencheck  & \greencheck  & \greencheck  & \greencheck  & \greencheck   & \textbf{38.23} & \textbf{25.25} \\
\bottomrule
\label{tab:ablation}
\end{tabular}
\vspace*{-0.1in}
\end{table}

\begin{table}[t!]
    \centering
    \caption{\blue{Rendering evaluation of multi agent fusion, voxel voting and optimization refinement.}}
    \begin{tabular}{@{}llccccc@{}}
    \toprule
    \textbf{Dataset} & \textbf{Method} & \textbf{PSNR $\uparrow$} & \textbf{SSIM $\uparrow$} & \textbf{LPIPS $\downarrow$} \\ \midrule
    \multirow{3}{*}{Campus Highbay} & Combine Maps & 21.10 & 0.575  & 0.514 \\
    & Voxel Voting & 21.87 & 0.572 & 0.518 \\
    & Refinement & \textbf{22.84} & \textbf{0.718} & \textbf{0.444} \\
    \midrule
    \multirow{3}{*}{Campus Outdoor} & Combine Maps & 15.76 & 0.421 & 0.619 \\
    & Voxel Voting & 16.58 & 0.423 & 0.600 \\
    & Refinement & \textbf{19.92} & \textbf{0.694} & \textbf{0.452} \\
    \bottomrule
    \label{tab:fusion}
    \end{tabular}
    \vspace*{-0.2in}
\end{table}

% --------------------------------------------------------------------------

\section{Discussion and Limitations}
\label{sec:discussion}

Task-conditioned allocation pays off in proportion to the asymmetry between the task region and the full scene: when the task is spatially concentrated, the matched-budget reallocation puts rendering fidelity where downstream perception will use it, and when the task covers most of the image, \ours{} degrades gracefully to near-uniform mapping with a small attention overhead. The Replica and Campus ablations (Tables~\ref{tab:ablation}) show that every component contributes a positive ROI PSNR gain on both synthetic and real data, with the largest single jumps from ROI seed budgeting and stratified initial scaling. The same task-allocation signal extends naturally to the multi-agent setting: because $\kappa_i$ records where each agent spent its active-phase budget, two task-specialized maps can be merged by a per-voxel vote without re-optimization, recovering near-specialist quality from a small ROI-only data exchange (Tab. \ref{tab:fusion}).

Several limitations are worth noting. Spatial predicate evaluation relies on rendered depth and may misfire when depth is severely noisy. The LLM parser can hallucinate target objects under unusual phrasings. We evaluate on a small set of task phrasings rather than comprehensively across natural-language variation. Finally, the current system does not re-allocate budget when the task changes mid-sequence: regions that were under-reconstructed cannot be retroactively refined without re-observing them or conducting further refinement offline.

% ============================================================================
% 6. CONCLUSION
% ============================================================================
\section{Conclusion}
\label{sec:conclusion}

We presented \ours, a task-driven 3DGS mapping system that conditions its representation density on a natural-language task.  By importing the principle of task-driven foveation from human vision (i.e. allocating representational resources not uniformly, but rather in proportion to task relevance), we achieve consistent ROI PSNR improvements over three strong baselines at matched Gaussian budget and mapping computation time. ROI-weighted seeding, small initial Gaussian scales, random keyframe-window sampling, and active-only optimization together let the system rendering quality surpass baselines in nearly all evaluated scene/task pairs.  We further demonstrated that two task-specialized agents can be fused via per-voxel active-optimization-count voting, preserving near-specialist quality in a single shared map. We believe this work opens a productive direction: treating 3D scene representations not as passive reconstructions, but as \emph{task-conditioned cognitive maps} whose fidelity follows the robot's purpose.

\bibliographystyle{IEEEtran}
\bibliography{references}

\end{document}